\documentclass[aip,amsmath,amssymb,reprint]{revtex4-2}

\usepackage{booktabs}
\usepackage[table]{xcolor}
\usepackage{graphicx}% Include figure files

\usepackage{hyperref} % add hypertext capabilities 
\usepackage{subcaption}
\usepackage{fancyhdr}

\fancypagestyle{firstpagefooter}{ \fancyhf{} % Clear all header and footer fields
    \fancyfoot[C]{Distribution Statement A.  Approved for public release: distribution is
    unlimited.} % Center header
}

\newcommand{\aref}[1]{\hyperref[#1]{Supplementary Information~\ref*{#1}}}
\makeatletter
\def\maketitle{
\@author@finish
\title@column\titleblock@produce
\suppressfloats[t]}
\makeatother

\begin{document}

\title{Symmetry constrained neural networks for detection \\and localization of damage
in metal plates}

\author{James Amarel} 
\email{jlamarel@lanl.gov} 
\affiliation{
 Los Alamos National Laboratory, Los Alamos, NM, 87545, USA}%

\author{Christopher Rudolf} 
\email{christopher.c.rudolf.civ@us.navy.mil}
\affiliation{% 
U.S. Naval
Research Laboratory, Washington, DC, 20375, USA }%

\author{Athanasios Iliopoulos} \affiliation{% 
U.S. Naval Research Laboratory, Washington, DC, 20375, USA }%

\author{John Michopoulos} \affiliation{% 
U.S. Naval Research Laboratory, Washington, DC, 20375, USA }%

\author{Leslie N. Smith} 
\affiliation{% 
U.S. Naval Research Laboratory, Washington, DC, 20375, USA }%

\date{\today}

\begin{abstract} 
The present paper is concerned with deep learning techniques applied to detection and
localization of damage in a thin aluminum plate.  We used data collected on a tabletop
apparatus by mounting to the plate four piezoelectric transducers, each of which took
turn to generate a Lamb wave that then traversed the region of interest before being
received by the remaining three sensors.  On training a neural network to analyze
time-series data of the material response, which displayed damage-reflective features
whenever the plate guided waves interacted with a contact load, we achieved a model that
detected with greater than $99\%$ accuracy in addition to a model that localized with
$2.58 \pm 0.12$ mm mean distance error. For each task, the best-performing model was
designed according to the inductive bias that our transducers were both similar and
arranged in a square pattern on a nearly uniform plate.
\end{abstract}

%\pacs{}% insert suggested PACS numbers in braces on next line

\maketitle %\maketitle must follow title, authors, abstract and \pacs

\section{Introduction} 
\label{sec:Introduction} 

Structural health monitoring (SHM) plays a crucial role in enhancing the
reliability, safety, and lifespan of critical structures and equipment,
including automatic drivelines \cite{DriveLine}, bridges \cite{Bridge},
concrete buildings \cite{Concrete}, and airplane wings \cite{AircraftPanel,
Aero}. Integrating sensors directly into system components facilitates
real-time monitoring, which allows for the early detection of potential issues,
such as cracks, corrosion, material loss, and delamination \cite{Farrar},
thereby extending service lifetimes.  Furthermore, minor problems can be
prevented from escalating into catastrophic failures by following maintenance
strategies prescribed according to the better-informed risk profiles made
available by intelligent sensing.

Nondestructive evaluation techniques are of particular interest for material
integrity determination.  In this respect, Lamb waves are especially desirable,
for they are able to travel long distances with low attenuation. Furthermore,
there exist standard procedures for their generation through the use of
specimen mounted piezoelectric transducers \cite{Cawley1996, Giurgiutiu2005,
Ostachowicz2009, Giurgiutiu2011}.  Moreover, Lamb waves feature multiple
vibrational modes, each of which offers varying sensitivity to different types of defects
\cite{RP2024b}.  

Through acoustic wave interrogation methods, previous workers have advanced
solutions via a number of signal-processing techniques, physical
considerations, and neural network designs.  The integration of deep learning
techniques into SHM routines has been shown to elevate the capabilities of
predictive maintenance algorithms \cite{Melville2018, Zhao2019, Malekloo2022,
OGW1, Azimi}.  These approaches excel relative to those based on classical
computing algorithms in the complex noisy environments common to real-world use
cases.

A hybrid data-driven/mechanistic approach for characterizing material defects
is that of physics-informed neural networks \cite{Karniadakis2022}.  Also
through the use of known physics, Zhang et al. \cite{Zhang2021} leveraged
calculated time-of-flight deviations together with a convolutional neural
network (CNN) to reconstruct an image indicating the probability of damage in
each of the possible defect locations.  For complex systems, however, it is
less clear how to employ approximate governing equations to deduce the location
and nature of damage from modifications to the response caused by wave-defect
interactions.  Indeed, effects due to specimen boundaries, manufacturing
processes, and sensor bonding are challenging to capture analytically  
\cite{Wang2022, Lemistre2001, Cui2022, Mitra2016, Barthorpe2010}.  Furthermore,
understanding the dispersive nature of Lamb waves, their temperature
dependence, and the nonlinear material response also demands significant time
investment from subject-matter experts if either physics models or manual
feature-extraction techniques are to be utilized \cite{Doebling1996}.  Of the
possible features to extract, a set that includes arrival times, amplitudes,
spectral coefficients, and time-correlation functions, it remains unclear how
to form the optimal descriptor. Alternatively, one can train deep learning
algorithms to implicitly extract the relevant damage indices directly from data
\cite{Yu2019, Guo2020}.

Rai et al. \cite{Rai2021} fed raw time-domain signals into a 1D CNN for
detection of notch-like damage. Xu et al.  \cite{Xu2024} located fatigue cracks
with a 1D-attention-CNN, using wavelet coefficients as features.  From
time-frequency representations, one can both detect anomalies \cite{Liu2019,
Khan2019} and localize cracks \cite{Rautela2021} using image-based techniques. 
Mariani et al. \cite{Mariani2021} found that using an adaptation of the causal
convolutional WaveNet architecture while forgoing conventional
baseline-subtraction methods afforded robust generalization capabilities  when
detecting defects in the presence of data distribution shift due to operating
temperature variations. Palanisamy et al.  \cite{RP2024a} investigated how to
transfer knowledge from a neural network trained on sparse sensor network data
from one component to another. It is an ongoing challenge to successfully
transition laboratory developed algorithms to their intended real-world use
cases due to factors including limited data availability, material and environmental
variability, sensor bonding inconsistencies, and the difficulty of generalizing across
different damage types, structures, and operating conditions \cite{Cawley2018}.  Song et al.
\cite{Song2023} utilized neural network
layers with both global and local context to localize damage in carbon
fiber-reinforced plastic laminate. They used  spherical damping soil placed
within a square region between four piezoelectric transducers as a proxy for
damage.  Use of a square array of transducers is common practice in the
literature \cite{Chenhui2019, Yelve2017}.

\begin{figure}
    \centering 
    \includegraphics[width=80mm]{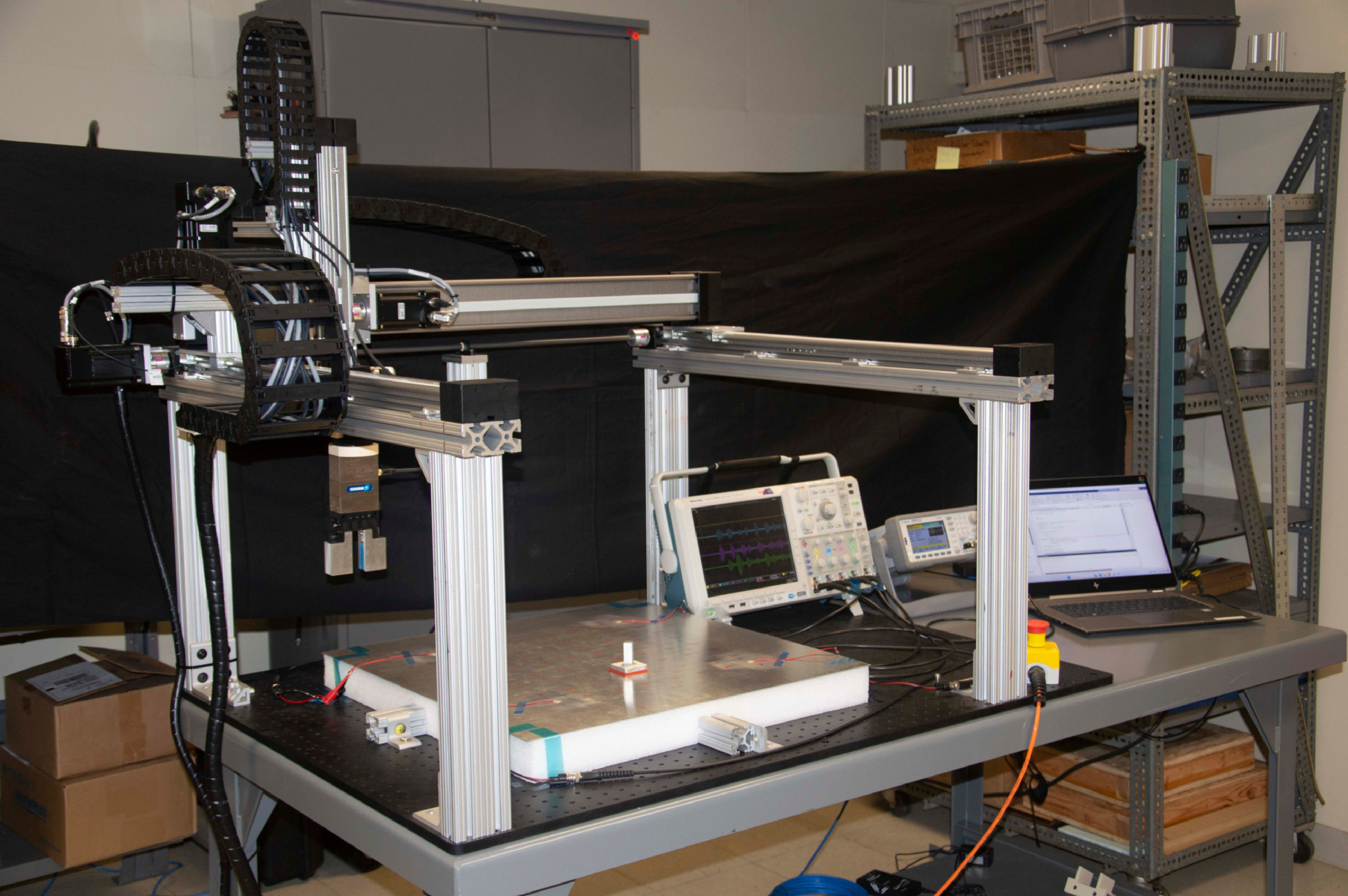}
    \caption{
Photo of the experimental apparatus showing the gantry, the aluminum
plate, the piezoelectric transducers, and the contact load in addition to the
waveform generating oscilloscope and a laptop that runs the code responsible
for controlling the gantry.
} 
\label{fig-apparatus} 
\end{figure}

In our experiment (see \autoref{fig-apparatus}), one transducer is placed near
each corner of a square plate and a contact load of square geometry is used as
a damage proxy. This sensor grid formally admits a geometric inductive bias,
which yields as constraints a set of necessary conditions on the neural network
architecture that in idealized conditions can be expected to increase per
parameter expressivity and bolster model generalizability \cite{GCN}; while one
cannot fully specify the architecture by such reasoning, it does follow from
these principled symmetry arguments that the multi-sensor data should be fused
by viewing the measured signals as living on a four-node graph. In this way,
the transducers live on nodes while the edges represent messages carried by
Lamb waves \cite{Zhou2022}. 

Real systems typically lack perfect symmetry, and the nature of the symmetry
breaking is often incompletely characterized. Consequently, it is not clear to
what extent one should hamper the flexibility of an ordinary neural network by
baking into the architecture a notion of equivariance.  Ideally, one could
retain the flexibility of ordinary neural networks while also realizing the
improvements to both robustness and generalizability that follow from symmetry
constrained modeling.

Even in the presence of exact symmetry, the problem of determining the optimal network
architecture remains unsolved. The situation is further complicated by the problem specific
nuances of symmetry breaking. That our results show a favorable outcome on using a slightly
modified form of the approximately equivariant structure proposed by Wang et al.
\cite{Yu2022} calls for further investigation into symmetry as an ingredient in
next-generation SHM routines.  Such research, together with interpretations of the learned
symmetry-breaking, could provide insights into the physical system, refine confidence
estimates of model predictions, and inform the development of techniques for achieving more
consistent performance when transferring trained models to different structures or
equipment.  Indeed, symmetry constraints alleviate data scarcity by reducing the effective solution space,
prevent overfitting to material and environmental noise by enforcing physically consistent
representations, and enhance adaptability for fine-tuning across different structures by
preserving fundamental geometric relationships.

In the following, we present the first study on the effects of incorporating
equivariance, both exact \cite{GCN} and approximate \cite{Yu2022}, associated
with sensor-network geometry into a deep learning algorithm tasked with
detecting and localizing damage from data reflecting acoustic wave propagation
in thin plates. In doing so, we exhibit the benefits of symmetry-aware modeling
and systematically address difficulties
reported by previous authors \cite{RP2024b, Shao2022} when measurements
corresponding to different damage states are related by symmetry transformations.  

This paper is organized as follows. In \autoref{sec:Data}, we describe and
visualize the data. Then, for the reader's convenience, we discuss our results
in \autoref{sec:Results} before considering the technical definitions of our
neural network architectures, which are given in \autoref{sec:Models}. Finally,
in \autoref{sec:Conclusion} we state our conclusion.

Our code makes use of the deep learning library Lux.jl \cite{Lux} and the plotting software
CairoMakie.jl \cite{CairoMakie}.

\section{Problem Statement} 
Structural health monitoring using Lamb waves is a widely studied approach for
detecting and localizing damage in thin-walled structures such as plates. Lamb waves, which
propagate as guided elastic waves in solid plates, are highly sensitive to defects, making
them suitable for non-destructive evaluation applications. The challenge, however,
lies in effectively extracting meaningful features from these wave signals and mapping them
to accurate damage predictions. The difficulty of this task is compounded by complex wave interactions,
including dispersion, mode conversion, and reflections from structural boundaries, which
obscure direct damage inference.

Our objective is
to infer either the existence of damage or the location $\bf{r}$ of existing damage
from time-series Lamb wave responses $\bf{s}$ measured at a finite set of
sensor locations.
Using supervised learning with an experimentally collected dataset $D$, we train the weights $\theta$ of a neural network
$F_{\theta}$ such that a cost function $E$ is minimized
\begin{equation}
    \label{cost}
    \theta^* = \arg\min_{\theta} \sum_i E[F_\theta(\bf{s}_i, \bf{r}_i)].
\end{equation}
Then, the (localization) mapping
\begin{equation}
    F_{\theta^*}:\bf{s} \mapsto \bf{r}
\end{equation}
can infer the location of damage from measurements $\bf{s}$ not seen during training,
thereby providing real-time, data-driven damage localization. The detection
mapping is similar, except rather than predicting $\bf{r}$, the network needs only to
solve a binary classification problem.

Identifying the optimal neural network architecture for wave-based damage
detection is difficult. The architecture must balance expressive
capacity, generalization ability, and computational efficiency while handling the spatial
and temporal complexities of Lamb wave propagation. Otherwise, the optimization algorithm
tasked with solving \autoref{cost} will not find an acceptable solution.

\section{Data} 
\label{sec:Data}

\begin{figure}
    \centering 
    \includegraphics[width=65mm]{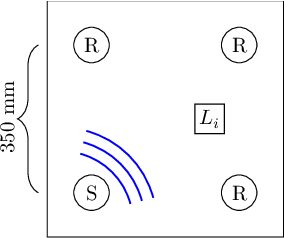} 
    \caption{Plate
schematic for Lamb wave source $S$ in the presence of a contact load at position $L_i$ that redirects
waves into receivers $R$.} 
\label{fig-plate} 
\end{figure}

Attached to an aluminum 6061-T6 plate of side length $610$ mm and thickness $1.2$ mm are
four APC International Model 63 piezoelectric transducers arranged in a square grid of
length $350$ mm on its sides.  A single $300$ kHz, $5$ count, Hanning-windowed tone burst
generates a vibrational waveform in one corner of the plate that then travels through the
bulk, thereby interacting with material inhomogeneities, before ultimately being registered
by three passive sensors in the flight path (see \autoref{fig-plate}). For each contact load
configuration, we then repeat this procedure to collect response measurements resulting from
the generation of a propagating wave in each corner.
As a result, we associate with every damage state a collection of sixteen
time-series signals, each of which contains $10,000$ piezoelectric voltage recordings.
Each of the $10,000$ recordings corresponds to voltage measurement at a particular instant in time. Recordings are taken at
equal intervals over a time window of $0.4$ ms following signal transmittance.

\begin{figure} 
    \centering 
    \includegraphics[width=65mm]{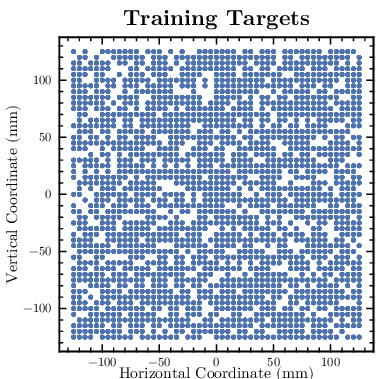} 
    \caption{A typical uniform initialization of a training set covering $80\%$ of the
    possible load locations. } 
\label{fig-traintest}
\end{figure}

In total, the dataset used in this paper comprises $2601$ damaged examples, each of which
corresponds to the contact load being placed in a different location within the square of
area $0.275$ $m^2$ concentric to the region bounded by our transducers, in addition to $6$
baseline examples where the contact load is absent. Half of the baseline examples were
determined at the beginning of data collection while the remaining three were measured after
all of the damaged examples had been acquired.  The contact load face, which is of a square
geometry with $40$ mm side length, couples to the plate via a $0.125$ inch thick silicone
sheet of $10$A durometer.  A $51 x 51$ square lattice (see \autoref{fig-traintest}) is
generated by placing the contact load at $5$ mm intervals.

Those locations in \autoref{fig-traintest} without a marker
are either in the test set or one of the
examples that we discarded. We rejected as poor quality $12$ examples
containing generated signals with spectral densities far from their intended form.
Additionally, we removed from our dataset
$8$ examples containing received signals that were uncharacteristically far, as measured
by Euclidean distance, to the appropriate mean received signal. When using the baseline-subtraction
technique, we also discarded $12$ examples that developed atypically large maximum signal
amplitudes.

\subsection{Compressed Signals}

\begin{figure} 
    \centering 
    \begin{subfigure}{0.5\textwidth}
        \includegraphics[width=85mm]{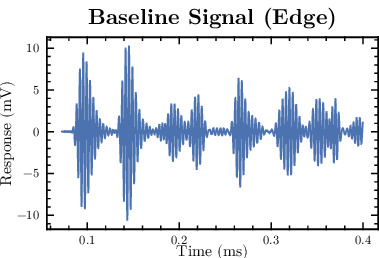}
        \caption{Baseline signal received across the edge path.}
        \label{subfig-edge_path}
    \end{subfigure} 
    \begin{subfigure}{0.5\textwidth}
        \includegraphics[width=85mm]{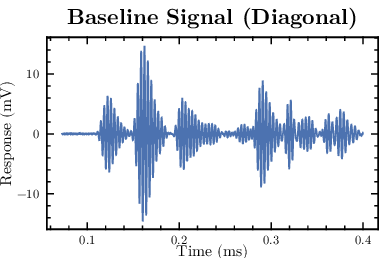}
        \caption{Baseline signal received across the diagonal path.}
        \label{subfig-diag_path}
    \end{subfigure} 
    \caption{Average full-fidelity received baseline signals for
        (\protect\subref{subfig-edge_path}) waves traversing
    along an edge connecting two transducers and (\protect\subref{subfig-diag_path}) waves
traversing the diagonal connecting two transducers.}
\label{fig-baseline_raw} 
\end{figure}

The raw received signals resolve details of wave propagation at fidelities unnecessary for
both localization and detection (see \autoref{fig-baseline_raw}). %
Compression of the signals can be achieved by examining their spectra. The dominant Fourier
modes are  contained within the $180kHz-420kHz$ interval \cite{Rautela2021b} (see
\autoref{fig-spectra_bl}). Our implementation of such a high and low pass filter yielded,
after backtransforming to real space, a time-series signal reduced from $10,000$ to $192$
elements in length. Then, discarding times before the signal had a chance to reach the
receivers, trimmed to length $158$ the time sequences.
\begin{figure*}
    \centering 
    \includegraphics[width=170mm]{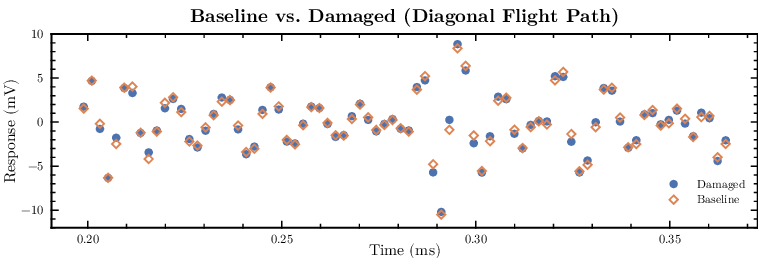}
    \caption{Comparison of a baseline signal that propagated along a plate diagonal in the
        absence of a contact load with a signal
    corresponding to a damaged state that followed the same flight path. Both signals
    have been compressed from their raw form by way a high and a
low pass filter.} 
\label{fig-pair_compressed} 
\end{figure*}

Having effectively reduced the sampling rate and filtered out high frequency noise, the
difference between baseline and damaged signals is readily visualized (see
\autoref{fig-pair_compressed}). Indeed, contact load induced attention is apparent
in the $0.20 - 0.24$ ms time window where the lowest-order antisymmetric longitudinal mode
$A_0$ is expected to arrive when traveling across the plate diagonal. Of second most
importance for localization are the $A_0$
waves that were reflected into a receiver by the specimen
boundary; for diagonal wavepaths, we recorded these signals in the time window $0.32 - 0.37$
ms. Though subdominant relative to the above two contributions, the symmetric modes (both
longitudinal and shear) that arrive before $A_0$ (see \autoref{fig-baseline_raw}) also carry
information pertaining to the material damage state.

\subsection{Structure of input data}
\begin{figure}
    \centering 
    \includegraphics[width=80mm]{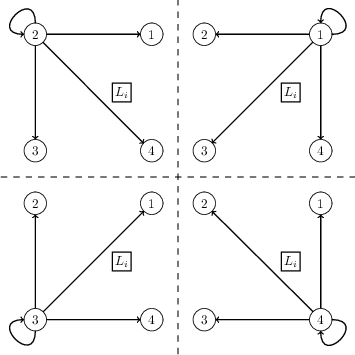} 
    \caption{Visual
    representation of the adjacency matrix structure, for an example with load location
    $L_i$. Diagonal elements correspond to self
interactions, while off-diagonal terms correspond to messages carried by Lamb waves across
the plate for a given receiver-sender pair.} 
\label{fig-adj_matrix} 
\end{figure}

To each damaged state, i.e., contact load center of mass vector $\mathbf{x}$, we associated an
adjacency matrix of sixteen time-series signals
$V_{rs}(t)$ of length $158$, one for each configuration specified by $r, s$, the receiver
and the sender index, respectively (see \autoref{fig-adj_matrix}).  This way, the measured
signals could be viewed as living
on a four-node graph, with transducers as the nodes and the edges corresponding to messages
carried by Lamb waves \cite{Zhou2022}. When the dihedral group is an exact symmetry, this
graph forms a homogeneous space that demands an equivariant architecture.

\section{Results} 
\label{sec:Results} 

Loss and accuracy evolution curves, in addition to cumulative error
distributions were obtained from results for model performance across $6$ different training
initializations, i.e., model parameters and train/test split. We investigate three model
types: 1) an \textit{ordinary} convolutional neural network, 2) an \textit{exactly
equivariant} neural network, and 3) an \textit{approximately equivariant} neural network.
These models were designed similarly in all respects except for their treatment of the square
group symmetry. Both symmetry-aware models possessed about $366,000$ trainable parameters,
while the \textit{ordinary} model operated with about $371,000$ trainable parameters.

For training, we used the Adam optimizer \cite{Adam} with batch size $32$ and the
OneCycle \cite{OneCycle} learning schedule defined by an initial learning rate of $10^{-5}$
that ramps for $200$ epochs until reaching strength $2.5 \times 10^{-3}$
before descending over the final $800$ training epochs. We used final learning rates
of $10^{-3}$ and $10^{-7}$ for the training of our locators and our detectors, respectively.

Only the \textit{exactly equivariant} model offered predictions constrained to 
transform under changes of coordinates generated by symmetry operations like vectors, in the
case of localization, or scalars, in the case of detection; however, the
\textit{approximately equivariant} model is initialized in an equivariant state and
possessed limited degrees of freedom for violating square symmetry (see
\autoref{ssec:approx}).

\subsection{Localization} 
\label{subsec:localization}

Toward optimizing model parameters, we first constructed the train/test datasets by
uniformly sampling locations to be held out at a $80\%/20\%$ ratio. Then, as a form of data
augmentation, we used baseline-subtraction to associate with each target location six
signals. For the training dataset, each baseline-subtracted example was treated individually
with respect to mini-batching. However, each test prediction was the result of averaging the
model output over all six baseline-subtracted inputs. This asymmetric treatment is expected
to equip the trained model with a statistical advantage against drift in operating
conditions during data acquisition \cite{Nerlikar2024}.  

\begin{figure}
    \centering
\includegraphics[width=85mm]{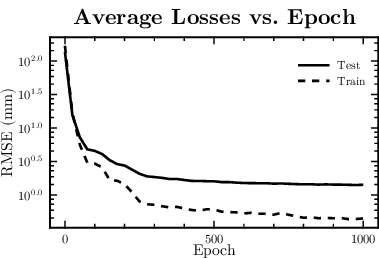} 
\caption{Test and train optimization loss evolution averaged across all
initializations and models. RMSE is the root mean square optimization error, \autoref{eq-L}.} 
    \label{fig-loss_avg}
\end{figure}

\begin{figure}
    \centering
\includegraphics[width=85mm]{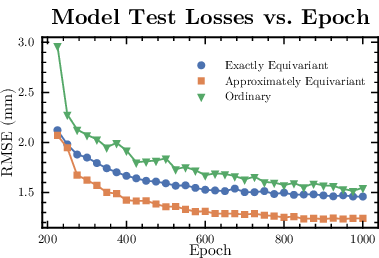} 
\caption{Comparison across models of the root mean square error (RMSE) evolution averaged across all
    initializations.} 
\label{fig-loss_compare_loss_md2e_iou_exc}
\end{figure}

Each model is trained (see \autoref{fig-loss_avg}) using the following pairwise loss
\begin{equation} 
    \text{E}(\hat{\mathbf{x}}, \mathbf{x}) = \text{max}(|\hat{\mathbf{x}} - \mathbf{x}|^2,
    |0.5\lambda_{A_0}|^2)
\big[1-A(\hat{\mathbf{x}}, \mathbf{x})\big],
\label{eq-L} 
\end{equation}
where $\hat{\mathbf{x}}$ and $\mathbf{x}$ point to the center of mass of the ground-truth
contact load and the predicted contact load, respectively and $A$ gives the percentage area
overlap between the predicted and ground-truth contact load faces.  Inspired
by image segmentation techniques, we designed a loss that
switches over to an intersection over union behavior for predictions with distance errors
less than the naive diffraction limit of one-half of the $A_0$ mode wavelength,
$\lambda_{A_0}  = 6.70$ mm. As a result of this choice, we expected, based on arguments 
purely statistical in nature, a reduction in severity of outlier
predictions, at the cost of an increase in mean error, relative to what would have been
achieved if mean squared error was used as the optimization loss function. Once the transient
features of training have subsided (see \autoref{fig-loss_compare_loss_md2e_iou_exc}), the \textit{approximately equivariant} model maintains a
markedly lower optimization loss on a hold-out test dataset
than both the \textit{exactly equivariant} model and the
\textit{ordinary} model. The \textit{approximately equivariant} model converges most quickly
and finds the superior function approximation.
\begin{table}
    \centering
    \resizebox{0.5\textwidth}{!}{  % Set table to 50% of text width
        \begin{tabular}{|c|c|c|c|}
            \hline
             & Exact (mm) & Approx. (mm) & Ordinary (mm)\\
            \hline
            MDE & $2.93 \pm 0.11$ & \cellcolor{lightgray}$2.58 \pm 0.12$ & $2.98 \pm 0.14$ \\
            \hline
            Var. & $1.84 \pm 0.12$ & \cellcolor{lightgray}$1.61 \pm 0.11$ & $1.92 \pm 0.18$\\
            \hline
            RMSE & $1.46 \pm 0.09$ & \cellcolor{lightgray} $1.24 \pm
            0.08$ & $1.55 \pm 0.17$ \\
            \hline
            Gap & $1.07 \pm 0.10$ & \cellcolor{lightgray} $0.84 \pm 0.09$ & $0.99 \pm 0.18$ \\
            \hline
        \end{tabular}
    }
    \caption{Test performance, as measured by the mean distance error (MDE),
    the variance of the distance errors (Var.), the root mean square error 
    (RMSE), and the generalization gap (Gap).}
    \label{tab-locate}
\end{table}

Out of the three models we studied, the \textit{approximately equivariant} model attained
the lowest mean values on all four metrics listed in \autoref{tab-locate}.  Evidently, the
weak constraints imposed upon the \textit{approximately equivariant} model typically
facilitated decreased variance and reduced generalization gap relative to both the strictly
constrained and unconstrained architectures. By allowing for weak symmetry breaking, we
achieved a better model with respect to metrics sensitive to both gross performance (MDE)
and worst case predictions (RMSE). 

\begin{figure}
    \centering 
    \includegraphics[width=85mm]{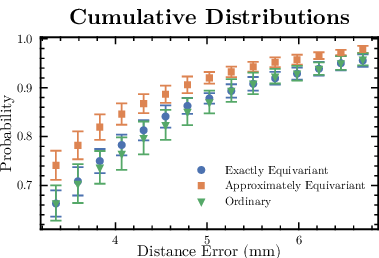}
    \caption{Comparison of truncated, initialization averaged, cumulative distributions on test data. All three models studied herein typically capture greater than $95\%$
of test data with errors less than $6.70$ mm.} 
\label{fig-cdf}
\end{figure}

Predictions by all three models studied herein were correct within $13.4$ mm
for $99.9\%$ of test data. The \textit{approximately equivariant} model
typically offered predictions within the diffraction limit on more than $70\%$ 
of the test examples (see \autoref{fig-cdf}). Only the \textit{ordinary} model admitted
test predictions of error greater than $16.75$ mm. 

Generically, the difficult examples are
those corresponding to load locations on the edges of the square region sampled
by our dataset (see \autoref{fig-hm_err} in \aref{app-results_loc}).
Boundary regions present substantial challenges in structural health monitoring due to
sparse sampling and the breakdown of translation equivariance. For damage locations near the
boundary of our sample grid, both the \textit{exactly equivariant} and the
\textit{approximately equivariant} architectures outperformed the
\textit{ordinary} model, yielding lower mean errors and reduced uncertainty (see
\autoref{fig-hm_err} and \autoref{fig-hm_flucts} in \aref{app-results_loc}).  This
improvement is attributed to the weight sharing mechanism inheret to group convolutions,
which facilitates the learning of features that generalize across symmetry related
examples. 

On evaluating the vector distance between transformed model predictions and model
predictions for transformed inputs, we found that the \textit{approximately equivariant}
model and the \textit{ordinary} model learned to violate equivariance on the boundary by
$80$ mm and $180$ mm, respectively (see \autoref{fig-hm_eqv_err} in \aref{app-results_loc}).
While some metrics (see \autoref{fig-fknn},
\autoref{fig-bl_eqv_err}, \autoref{fig-input_eqv_err}, and \autoref{fig-spectra_bl} in
\aref{app-data_eqv}) suggest that square symmetry is not weakly broken in our system, inspection
of the symmetry-breaking weights reveals that the approximately equivariant model
learned latent feature maps that exhibited diminishing equivariance violations with
increasing layer depth (see \autoref{fig-sym_weights} in \aref{app-results_loc}). In the
final layer, the symmetry-breaking weights deviated by less than $6\%$ from their
symmetry-preserving form.

\subsubsection{Time-Window Ablation}
The input data exhibits features
reflecting the multi-mode composition of the plate's vibrational spectrum together with
interactions between directly received waves and those that are redirected by the plate
boundaries.   
Although one might anticipate beneficial effects would follow from selectively
windowing the directly received lowest-order antisymmetric longitudinal mode $A_0$, as these
excitations tend to leak most strongly into contact loads, we empirically concluded  that
this act of manual feature selection is detrimental to model performance.
\begin{table}
    \centering
    \resizebox{0.5\textwidth}{!}{  % Set table to 50% of text width
        \begin{tabular}{|c|c|}
            \hline
             Time Window (ms)& Mean Distance Error (mm) \\
            \hline
             \cellcolor{lightgray}$0.07$ - $0.40$&\cellcolor{lightgray} $2.81 \pm 0.12$\\
            \hline
            $0.16$ - $0.40$& $3.05 \pm 0.21$\\
            \hline
            $0.07$ - $0.24$& $7.41 \pm 0.43$\\
            \hline
            $0.16$ - $0.24$& $10.39 \pm 0.71$\\
            \hline
        \end{tabular}
    }
    \caption{Comparison of mean model performance when ingesting the full compressed signal
    ($0.07$ - $0.40$ ms), vs. discarding the early to arrive symmetric mode ($0.16$ - $0.40$
ms), vs. truncating wave reflections ($0.07$ - $0.24$ ms), vs. windowing only the directly
received $A_0$ mode ($0.16$ - $0.24$ ms).}
    \label{tab-locate_abl}
\end{table}
Models that
ingested the entire signal after data curation consistently performed better than those that
either windowed the $A_0$ mode, truncated signal regions containing reflections, or
discarded the early to arrive symmetric modes (see \autoref{tab-locate_abl}).
Evidently, neural network based structural health monitoring solutions do not
require extensive physical modeling in order to glean useful information from measurements of
multi-mode acoustic excitations that underwent scattering events before arriving at a
transducer \cite{RP2024b}.

\subsection{Detection} 
\label{sec:Detection}

\begin{figure*}
    \centering 
    \includegraphics[width=170mm]{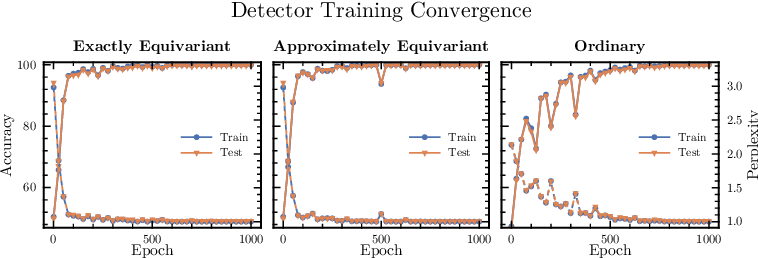}
    \caption{Accuracy and loss vs. epoch curves averaged over initializations for each
    architecture.} 
\label{fig-acc} 
\end{figure*}

In order to train a detector, we first addressed the class imbalance of our dataset by
forming synthetic normalized linear combinations of the six raw baseline examples until the damaged and
undamaged pairs were equal in number \cite{SMOTE}. Binary cross-entropy served as our
optimization loss (see \autoref{fig-acc}).

\begin{figure}
    \centering 
    \includegraphics[width=85mm]{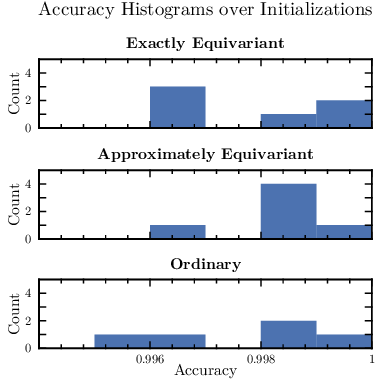}
    \caption{Histograms of final detector accuracies over distinct training
    initializations.} 
\label{fig-acc_hist} 
\end{figure}

For all three models studied herein, we achieved over $99\%$ mean accuracy on averaging
across the different initializations when using a $20\%/80\%$ ratio for our train/test
split. Final average training accuracies were $0.998 \pm 0.001$, $0.998 \pm 0.001$, and
$0.998 \pm 0.002$, for the \textit{exactly equivariant}, \textit{approximately equivariant},
and \textit{ordinary} models, respectively (see \autoref{fig-acc_hist}). Throughout this
subsection, error bars are computed as the difference between the mean value of a quantity
and its observed value nearest unity.  With perplexity defined here as the natural
exponential of the cross-entropy, we observed final perplexity values of $1.009 \pm 0.004$,
$1.008 \pm 0.004$, and $1.009 \pm 0.006$ for the \textit{exactly equivariant},
\textit{approximately equivariant}, and \textit{ordinary} architectures, respectively. 

Though all three models converged to states offering similar performance,
the symmetry-aware models required fewer training epochs 
than the \textit{ordinary} model.
This may be
because attenuation is recognizable (see \autoref{fig-pair_compressed}) with less regard
for those spatial inhomogeneities affecting arrival times that must be learned by the
localizer models.

\section{Models} 
\label{sec:Models}
%====================================================================
We studied three neural networks trained to recognize signatures of the
interactions between vibrational waves propagating through a thin-plate
specimen and a contact load, which is to be either detected or located, that
tends to absorb incoming wave energy and thereby serves as a proxy for
corrosion damage \cite{Muller2017}.  This task can be viewed as an inverse problem in which one
seeks to infer the state of a material from response measurements.

To this end, we first trained a $6$-block \textit{ordinary} convolutional
neural network in which the first five blocks contained a convolutional layer
of kernel size equal to the input sequence length that was symmetrically padded
such that the outgoing sequence length became half that of the input sequence
length.  The receiver and sender indices (see \autoref{fig-adj_matrix}) were
treated as channels.  Nonlinearities were omitted in the first block, which
used a length $158$ kernel. Internal layers included a skip connection followed
by LayerNorm \cite{layernorm} and the \textit{swish} \cite{Swish} activation
function.  In order to reduce internal layer inputs to the same size as their
output, as required by the skip connection, we convolved against a
non-trainable averaging kernel with the same padding and size as the associated
trainable convolution.  For the last block, where boundary effects were present
in the time-series, a dense layer acted on both channel and time indices. Then,
activation by \textit{tanh} preceded a final dense layer that yielded an output
array with two components, which was taken to contain the horizontal and
vertical coordinates of the load location.  Our models were regularized by
weight decay value of strength $10^{-6}$ and $0.05$ probability of dropout on
the channel index. 

On considering symmetries as a guiding principle for model design, it follows that there are
a number of constraints on the architecture that must be satisfied. These priors demand that
the trainable weights be shared in a manner consistent with time-translation invariance, the
arrow of time, and the square group \cite{GCN}.  In idealized conditions, this is expected to
maximize the expressivity of each neuron, ultimately yielding a model that is better suited
to perform on noisy data. 

Time-translation invariance, modulo boundary effects, and the arrow of time are readily
enforced by using layers with the standard convolutional action on the time index. The use
of varying lead lengths for transducer-oscilloscope communication further motivated 
convolutional architectures, since invariance to time-shifts is desired. Because only affine
layers admit shift equivariance \cite{Yu2021}, the arbitrariness of our choosing a reference
voltage is reflected only by omitting zero-frequency modes from the signal data. The scale
invariance required by our freedom in choice of units can be satisfied by appropriately
applying LayerNorm.  Lastly, owing to the square arrangement of the piezoelectric sensors,
it remains to incorporate equivariance with respect to symmetry transformations in the
square group. 

An \textit{exactly equivariant} architecture assumes that the only significant
spatial structure in our data is that of the sensor geometry and the load location;
however, this idealization is not borne out by the data.  Hence, we also put
forth an \textit{approximately equivariant} architecture that possesses a small
fraction of symmetry-breaking weights for the purpose of capturing those
aspects of both the plate and the transducers that produce deviations from
system homogeneity even in the absence of damage (see \autoref{fig-bl_eqv_err} in
\aref{app-data_eqv}).

\subsection{Dihedral group convolution}
\label{sec:dihedral}
%====================================================================

\begin{figure} 
    \centering 
    \includegraphics[width=80mm]{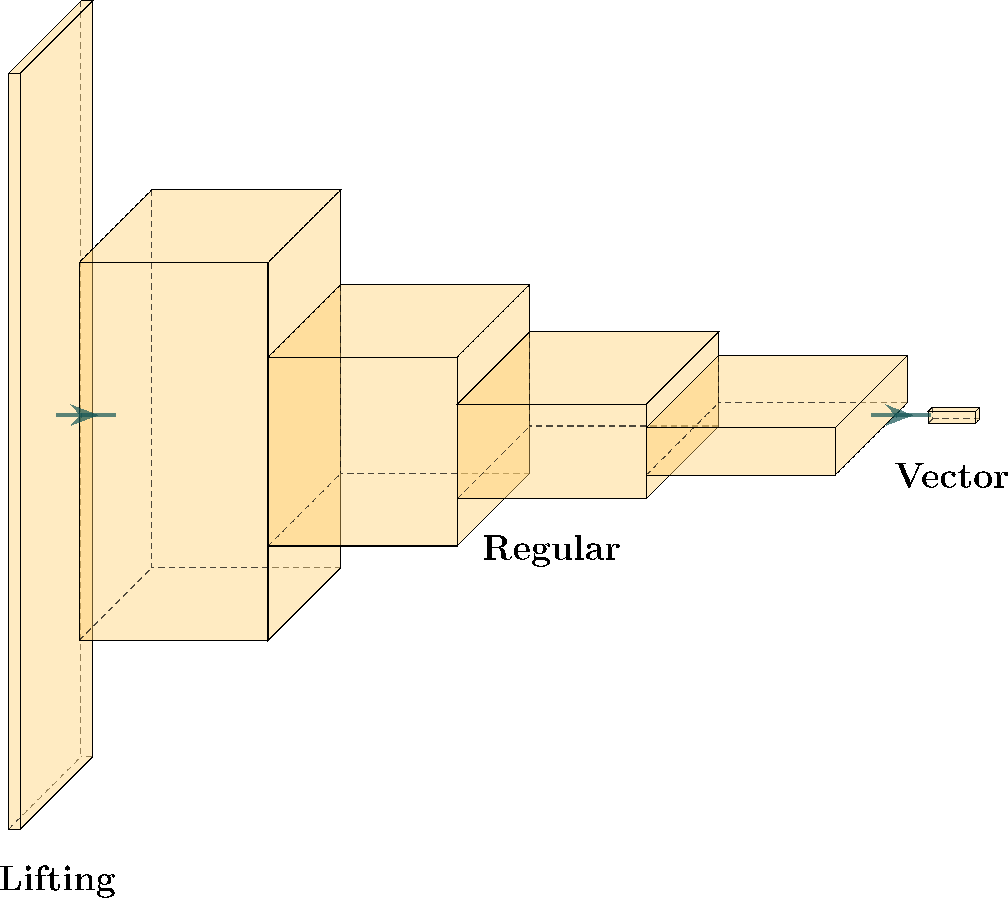} 
    \caption{Schematic of the model architecture \cite{PNN}. Each of the lifting
        and regular blocks contain a convolutional layer. Height indicates
        time-sequence length, depth reflects the channel number, and width
        corresponds to the group index. }
        \label{fig-architecture} 
    \end{figure}

Here, we view the features $V_{sr} = V(e_s, e_r)$ as a bilinear map defined by its action on
the canonical basis vectors $e$, where $r, s \in \mathbb{Z}_4$ correspond to the receiver
and sender transducer labels, respectively. On noting that $r, s$ transform jointly under permutations
$\sigma \in D_4$, with $D_4$ the square group, it follows that the linear action of the
first, lifting, layer of the group convolutional neural network must take the form
\cite{GCN}
\begin{equation} 
    \tilde{V}_{\sigma} = \sum_{r,s \in \mathbb{Z}_4}\,K(\sigma^{-1}e_s,
\sigma^{-1}e_r)V_{rs}, 
\label{eq-first_layer} 
\end{equation}
where $\sigma$ is realized in its four-dimensional representation and $K$ is the convolutional
filter.   Note that the time arguments of $V_{rs}(t)$ are suppressed in this section. 
Together with summing over the base space $\mathbb{Z}_4 \otimes \mathbb{Z}_4$ in
\autoref{eq-first_layer}, we include a time convolution, for the relevant group to our
system is that of time translations, cyclic rotations, and roto-reflections.
With $\tilde{V}$ a function on $D_4$, we use the right regular representation for
our internal layers, as this renders admissible pointwise nonlinearities
\cite{GeoDL}.  The internal linear actions of the network can then be
expressed
\begin{equation} 
    \hat{V}_{\sigma} = \sum_{\pi \in D_4}\,K(\sigma^{-1}\pi)\tilde{V}_{\pi}.
\label{eq-inner_layer} 
\end{equation}
To obtain a vector output $\mathbf{v}$ with components $v^i$ from features $\hat{V}_{\sigma}$
in the regular representation, we contract the rotation matrices $R^{ij}_{\sigma}$ along
both the group index $\sigma$ and a channel index $j$ of dimension two according to
\begin{equation} 
    v^i = \sum_{\sigma \in D_4} \sum_{j \in \mathbb{Z}_2}
R^{ij}_{\sigma}\hat{V}^j_{\sigma}.  
\label{eq-vector} 
\end{equation}
The above development is used in the construction of our \textit{exactly equivariant} model
(see \autoref{fig-architecture}).

        We note that an equation analogous to \autoref{eq-vector} can be used to output a
        p4m-equivariant \cite{GCN} grid suitable for probability imaging.

        \subsection{Approximately equivariant convolution}
        \label{ssec:approx}
        A complicating feature of our specimen is its inherent anisotropy caused by the
        rolling process used in manufacturing, which affects wave propagation and implies
        that the measured response is not independent of transducer location.  Similarly,
        effects due to boundary irregularities and bulk material inhomogeneities violate
        square symmetry. Further symmetry-breaking effects stem
        from non-ideal aspects of Lamb wave generation.

        If the symmetry-breaking is sufficiently weak, then incorporating a notion of
        approximate equivariance remains well motivated. This can be achieved in a manner
        that retains the structure of a strictly constrained neural network by introducing
        arrays containing eight symmetry-breaking trainable weights
        \begin{equation} \omega(g) = 8 \times \text{softmax}(g / 8), \label{eq-w}
        \end{equation}
        and augmenting the group convolution \cite{Yu2022}
        \begin{equation} \hat{V}_{\sigma} \rightarrow \omega(\sigma)\sum_{\pi \in
        D_4}K(\sigma^{-1}\pi)\tilde{V}_{\pi}.  \label{eq-appx} \end{equation}
        The weights in \autoref{eq-w} can be analogously introduced to
        \autoref{eq-first_layer} and \autoref{eq-vector}.  Note that \autoref{eq-w} is
        chosen to ensure that the elements of $\omega$ are positive definite, bounded from
        above, and do not introduce a superfluous trainable scale degree of freedom.  It is
        through \autoref{eq-appx} that we constructed our \textit{approximately equivariant}
        model.

\section{Conclusion} 
\label{sec:Conclusion}
%--------------------------------------------------------------------
We compared three neural network architectures on their ability to both detect and
infer the location of contact loads placed on an aluminum plate from time-series
measurements of the material response.  Two of these
architectures were designed to be aware of the geometry associated with our sensor
network, while the third architecture was not equipped with equivariance as an
inductive bias.

In the presence of symmetry breaking effects due to both material
irregularities and imperfect sensor function, we designed an approximately
equivariant neural network that achieved a mean distance error of $2.58 \pm
0.12$ mm. Tasked with the same problem, our ordinary and exactly equivariant
models attained $2.98 \pm 0.14$ mm and $2.93 \pm 0.11$ mm mean distance error,
respectively. The approximately equivariant model also reached the lowest
optimization loss. Our loss function favors reducing outliers over
maximizing mean performance, leading to a model that is more robust and less
susceptible to highly erroneous predictions.

By capturing $99\%$ percent of test examples with errors less than $10.05$ mm,
all three of our models, together with our dataset, constitute
state-of-the-art results with respect to localizing damage using machine learning techniques
together with Lamb wave sensing. 

When detecting the presence of a contact load on the plate, we attained
over $99\%$ accuracy with each of the three models studied herein. The symmetry
aware models required fewer training epochs to reach this level of performance. 

Our work demonstrates that equivariance constraints are beneficial in the complex and noisy
scenarios that can be expected in real-world
structural health monitoring applications.  Future
research may include using contact loads of varying geometry across multiple temperatures in
addition to the localization and detection of crack-like damage.

\section{Limitations}
While our study demonstrates the benefits of approximate equivariance,
several limitations must be acknowledged.

First, our training and testing data were collected using a single square plate. This setup
may introduce biases related to the specific material properties and geometry. While our
methodology based on symmetry is expected to generalize across different structures, further
investigation is necessary to quantify robustness against variations in transducer
placement, material properties, and material geometry. Second, we did not systematically
vary environmental factors such as temperature, humidity, and background vibrations.  Third,
our localization loss function was designed to optimize both distance accuracy and damage
region overlap, while also accounting for wavelength-based resolution limits. While this
loss function improves localization performance  on our dataset, an alternative formulation
would be needed if damage proxies were not limited to a single type. Future studies should
investigate the impact of varying damage size and character.

Despite these limitations, the demonstrated improvements in localization
performance due to approximate equivariance suggest that symmetry-aware models
have potential advantages in practical damage detection applications across a
broad range of structural configurations.
\section{Supplementary Material}
We include a supplementary visualizations to support and elaborate on the main text.
This material includes an investigation into sources of symmetry breaking in our system,
additional model performance and uncertainty figures, in addition to an inspection of the
trained symmetry breaking weight values.

\begin{acknowledgments} 
JA was supported by an NRC Research Associateship Program
at the U.S. Naval Research Laboratory (NRL), administered by the Fellowships Office of
the National Academies of Sciences, Engineering, and Medicine. CR, AI, and JM
acknowledge the support by the Office of Naval Research via NRL’s core funding.

JA thanks Gerd J. Kunde for fruitful discussions.

\end{acknowledgments}

\section*{Author Declarations}
\subsection*{Conflict of Interest}
The authors have no conflicts of interest to disclose.
\subsection*{Author Contributions}
\textbf{JA}: Conceptualization (lead), Data Curation (equal), Formal Analysis
(lead), Methodology (lead), Software (lead), Validation (lead), Visualization
(lead), Writing/Original Draft Preparation (lead), Writing - review and editing
(equal). 
\textbf{CR}: Conceptualization (supporting), Data Curation (equal), Funding acquisition (lead),
Investigation (supporting), Writing – review and editing (equal), Resources (equal).
\textbf{AI}: Conceptualization (supporting), Data curation (equal), Methodology
(supporting), Writing - review and editing (equal), Investigation (support),
Resources (equal), Software (support), Validation (support).
\textbf{JM}: Conceptualization (supporting), Methodology (supporting),
Writing-review and editing (equal), Investigation (support),
Resources (equal), Software (support), Validation (support).
\textbf{LNS}: Conceptualization (supporting), Methodology (supporting),
Writing - review and editing (equal), Resources (equal).
\section*{Data Availability}
The code and compressed data needed to reproduce the findings of
this study are openly available at \url{https://github.com/USNavalResearchLaboratory/MINC/}.
Raw data is available from either corresponding author upon reasonable request.

% \bigskip
% \hrule

\bigskip
\hrule
% Create the reference section using BibTeX:
\bibliography{main}

\clearpage
\title{Supplementary Information: Symmetry constrained neural networks for detection and localization of damage
in metal plates}

\maketitle
% Add supplemental to main document to prevent hyperref breaking due to a pdfLaTeX bug
\onecolumngrid

\renewcommand{\thefigure}{S\arabic{figure}}
\setcounter{figure}{0}
\renewcommand{\theequation}{S\arabic{equation}}
\setcounter{equation}{0}
\setcounter{section}{0}
\setcounter{page}{1}
% Undoes PRL default style for omitting section numbers (for arXiv only)
\setcounter{secnumdepth}{2}

\section{Measures of System Symmetry}
\label{app-data_eqv}
\begin{figure}[p]
    \centering 
    \includegraphics[width=170mm]{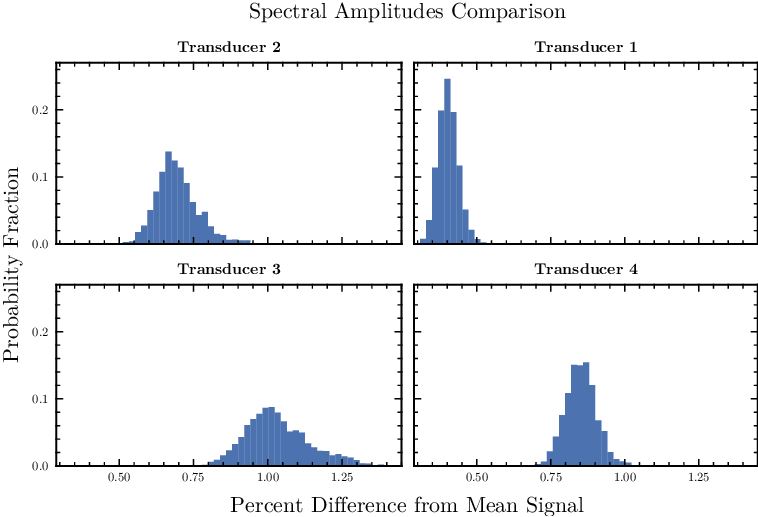}
\caption{Truncated histograms of the Euclidean distance between a given signal's spectral
amplitude and the mean spectral amplitude. Rejected examples are not shown.} 
\label{fig-fknn}
\end{figure}

One source of symmetry breaking in our system was the result of differences among the
transducer generated waves, which exhibited variations in amplitude, phase, and spectra. On
comparing the spectral amplitude of source waveforms (see \autoref{fig-fknn}), one sees that
while each transducer is of a distinguishable character, all generated waves remain within
$1.5\%$ of their typical form. We empirically concluded that these variations in the
generated waveforms were not a dominant source of difficulty for our models by selectively
investigating prediction errors corresponding to signals generated by transducer $3$ of
spectral amplitude that differed in excess of $1.25\%$ from the mean spectral amplitude. 

\begin{figure}
    \centering
    \includegraphics[width=170mm]{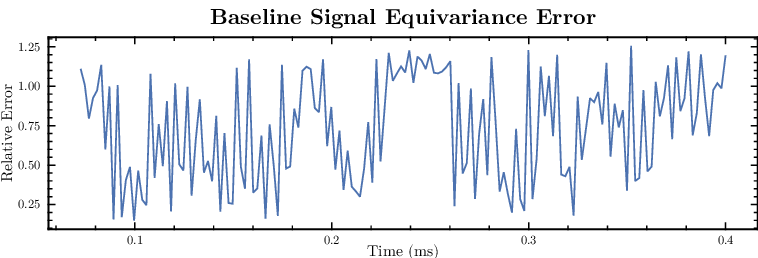}
    \caption{Baseline received signal equivariance error.}
    \label{fig-bl_eqv_err}
\end{figure}

\begin{figure}[p]
    \centering
    \includegraphics[width=170mm]{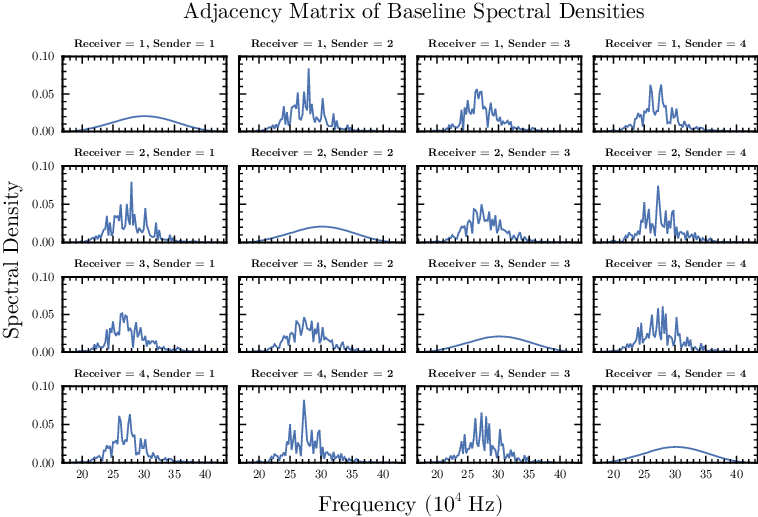}
    \caption{Average Fourier spectral densities of the baseline signals.}
    \label{fig-spectra_bl} 
\end{figure}

Contributions to symmetry-breaking in the absence
of damage include the influence of material anisotropies on wave propagation. These effects
are reflected in the baseline signals $V_{rs}^{(0)}(t)$, which would be invariant under
the square group if symmetry was exact. Thus, a measure of the bare
equivariance breaking in our system is provided by the normalized error
\begin{equation}
    R^{(0)}(t) = \frac{1}{8}\sum_{g\in G}\frac{\|V^{(0)}(t) - \rho_g
        V^{(0)}(t)\|}{\|V^{(0)}(t)\|}
\end{equation}
where $\rho_g$ is the appropriate matrix representation of an element $g$ in the square
group. On averaging over our $6$ baseline signals, in addition to neglecting both the
diagonals of $V_{rs}$ and times $t$ before the first Lamb wave arrival, we found $R^{(0)} =
0.73 \pm 0.33$ for the time-averaged relative equivariance error values (see
\autoref{fig-bl_eqv_err}). 
Symmetry-violating features in the baseline signals persist even
when neglecting both phase and scale differences (see \autoref{fig-spectra_bl}).

\begin{figure}
    \centering 
        \includegraphics[width=85mm]{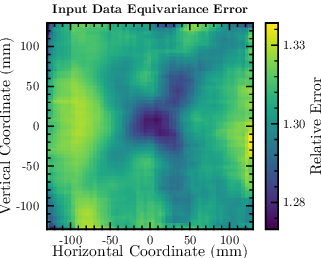}
    \caption{Heatmap of the input data equivariance error.}
    \label{fig-input_eqv_err}
\end{figure}

Visualization of the training input data symmetry breaking can be achieved by considering
the normalized error field
\begin{equation}
    R(\mathbf{x}) = \frac{1}{8}\sum_{g\in G}\frac{\|V(\mathbf{x}) - \rho_g
        V(\rho_{g^{-1}}\mathbf{x})\|}{\frac{1}{2}(\|V(\mathbf{x})\| + \|V(\rho_{g^{-1}}\mathbf{x})\|)}
\end{equation}
with $V(\mathbf{x})$ the adjacency matrix of baseline subtracted signals paired with the
target location $\mathbf{x}$, where the contact load is placed (see
\autoref{fig-input_eqv_err}). By this measure, symmetry violations are strongest near the
boundaries of our dataset, and weakest near the center of the plate. 

Throughout, heatmaps are constructed, using a $51 \times 51$ grid of pixels $5$ mm on each side,
as follows. For each contact load location, associated field data is attributed
to the pixel that is concentric with the load in addition to all pixels
completely covered by the physical extent of the contact load. We then average over all nonzero contributions to each pixel.
\section{Additional Locator Visualizations}
\label{app-results_loc}
\begin{figure}
    \centering 
        \includegraphics[width=170mm]{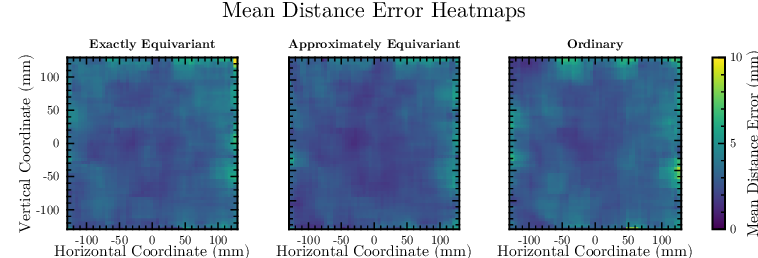}
    \caption{Initialization averaged mean distance error heatmaps.} 
\label{fig-hm_err} 
\end{figure}

\begin{figure}
    \centering 
        \includegraphics[width=170mm]{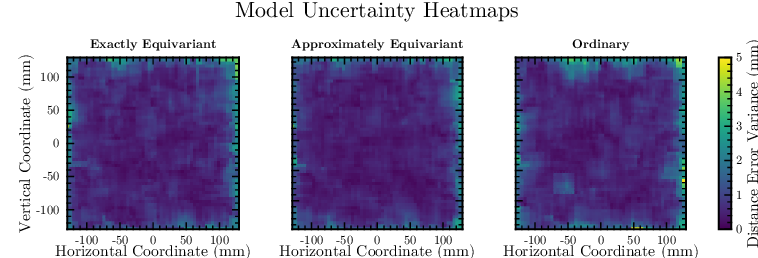}
    \caption{Inference error variance heatmaps.} 
\label{fig-hm_flucts} 
\end{figure}

\begin{figure}
    \centering 
        \includegraphics[width=170mm]{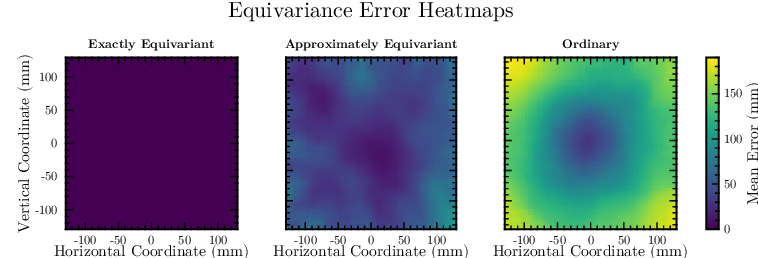}
    \caption{Initialization averaged learned equivariance error heatmaps.} 
\label{fig-hm_eqv_err} 
\end{figure}

\begin{figure}
    \centering
    \includegraphics[width=170mm]{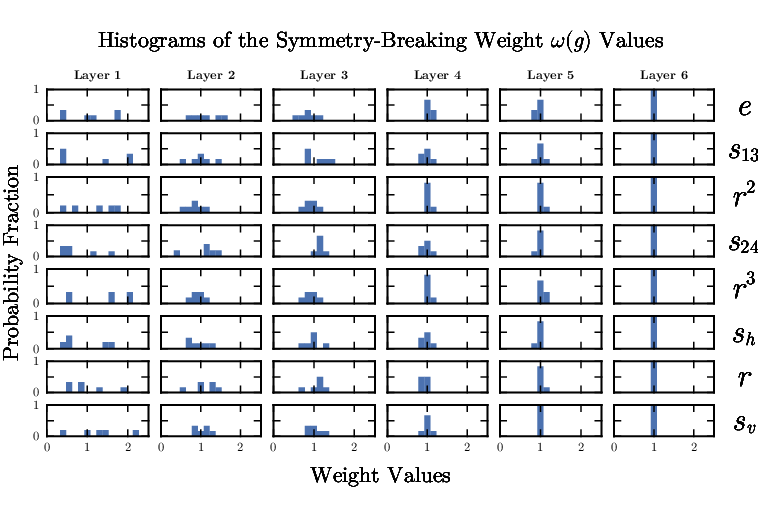}
    \caption{
Layer-wise distribution of the \textit{approximately equivariant} model's learned
symmetry-breaking weights. Here, the group elements are indicated by $e$ for the
identity element, $r$ for a $\pi/2$ rotation, $s_v$ for a reflection across the
vertical axis, $s_h$ for a reflection across the horizontal axis, and $s_{13}$ and
$s_{24}$ for reflections across the diagonal connecting corner $1$ with $3$ and $2$
with $4$, respectively.
}
    \label{fig-sym_weights}
\end{figure}

Expected performance can be visualized by averaging the test errors
across different initializations (see \autoref{fig-hm_err}). 
Similarly, we also
estimate model uncertainty by calculating the pointwise error variance (see
\autoref{fig-hm_flucts}). 
Additionally, a heatmap of the learned equivariance
error can be obtained as follows (see \autoref{fig-hm_eqv_err}). Let $\Psi$
be a neural network designed to act on $V$, an adjacency matrix
of signals. The mean equivariance error field is then
\begin{equation}
	Q(\mathbf{x}) = \frac{1}{8}\sum_{g\in G}\|\rho_g \Psi[V(\mathbf{x})] - \Psi[\rho_g V(\mathbf{x})]\|.
\end{equation}
While there is limited similarity between the learned equivariance errors (see
\autoref{fig-hm_eqv_err}) and our naive measure of input data equivariance (see
\autoref{fig-input_eqv_err}), the \textit{approximately equivariant} model
appears to learn some nontrivial structure. Evidence for this assertion lies in the fact
that the \textit{approximately equivariant} model learns symmetry-breaking weights 
that in the deeper layers only weakly deviate from unity (see \autoref{fig-sym_weights}).
\end{document}